\documentclass[letterpaper]{article} 
\usepackage{aaai25}  
\usepackage{times}  
\usepackage{helvet}  
\usepackage{courier}  
\usepackage[hyphens]{url}  
\usepackage{graphicx} 
\usepackage{natbib}  
\usepackage{caption} 
\usepackage{algorithm}
\usepackage{algorithmic}

\usepackage{amsmath,amsfonts}
\usepackage{algorithmic}
\usepackage{array}
\usepackage[caption=false,font=normalsize,labelfont=sf,textfont=sf]{subfig}
\usepackage{textcomp}
\usepackage{url}
\usepackage{verbatim}
\usepackage{graphicx}
\usepackage{multirow}
\usepackage{amsmath}
\usepackage{xspace}
\newcommand{\ie}{{\emph{i.e.}}\xspace}
\newcommand{\eg}{{\emph{e.g.}},\xspace}
\newcommand{\etc}{\emph{etc}}

\usepackage{booktabs}
\usepackage{colortbl}  
\usepackage{xcolor}
\usepackage{newfloat}
\usepackage{listings}
\DeclareCaptionStyle{ruled}{labelfont=normalfont,labelsep=colon,strut=off} 
\floatstyle{ruled}
\newfloat{listing}{tb}{lst}{}
\floatname{listing}{Listing}
\title{VIoTGPT: Learning to Schedule Vision Tools in LLMs \\ towards Intelligent Video Internet of Things}
\author{
    Yaoyao Zhong,
    Mengshi Qi,
    Rui Wang,
    Yuhan Qiu, 
    Yang Zhang, 
    Huadong Ma\thanks{Corresponding author: Huadong Ma.}
}
\affiliations{
    \textsuperscript{\rm 1}State Key Laboratory of Networking and Switching Technology, \\ Beijing University of Posts and Telecommunications, China\\
    \{zhongyaoyao, qms, mhd\}@bupt.edu.cn

}

\usepackage{bibentry}

\begin{document}

\maketitle

\begin{abstract}
Video Internet of Things (VIoT) has shown full potential in collecting an unprecedented volume of video data. How to schedule the domain-specific perceiving models and analyze the collected videos uniformly, efficiently, and especially intelligently to accomplish complicated tasks is challenging. To address the challenge, we build VIoTGPT, the framework based on LLMs to correctly interact with humans, query knowledge videos, and invoke vision models to analyze multimedia data collaboratively. To support VIoTGPT and related future works, we meticulously crafted the VIoT-Tool dataset, including the training dataset and the benchmark involving 11 representative vision models across three categories based on semi-automatic annotations. To guide LLM to act as the intelligent agent towards intelligent VIoT, we resort to the ReAct instruction tuning method based on VIoT-Tool to learn the tool capability. Quantitative and qualitative experiments and analyses demonstrate the effectiveness of VIoTGPT. We believe VIoTGPT contributes to improving human-centered experiences in VIoT applications. The project website is \url{https://github.com/zhongyy/VIoTGPT}.
\end{abstract}

\section{Introduction}
\label{sec:intro}

\begin{figure*}[htbp]
	\center
    \includegraphics[width=1\textwidth]{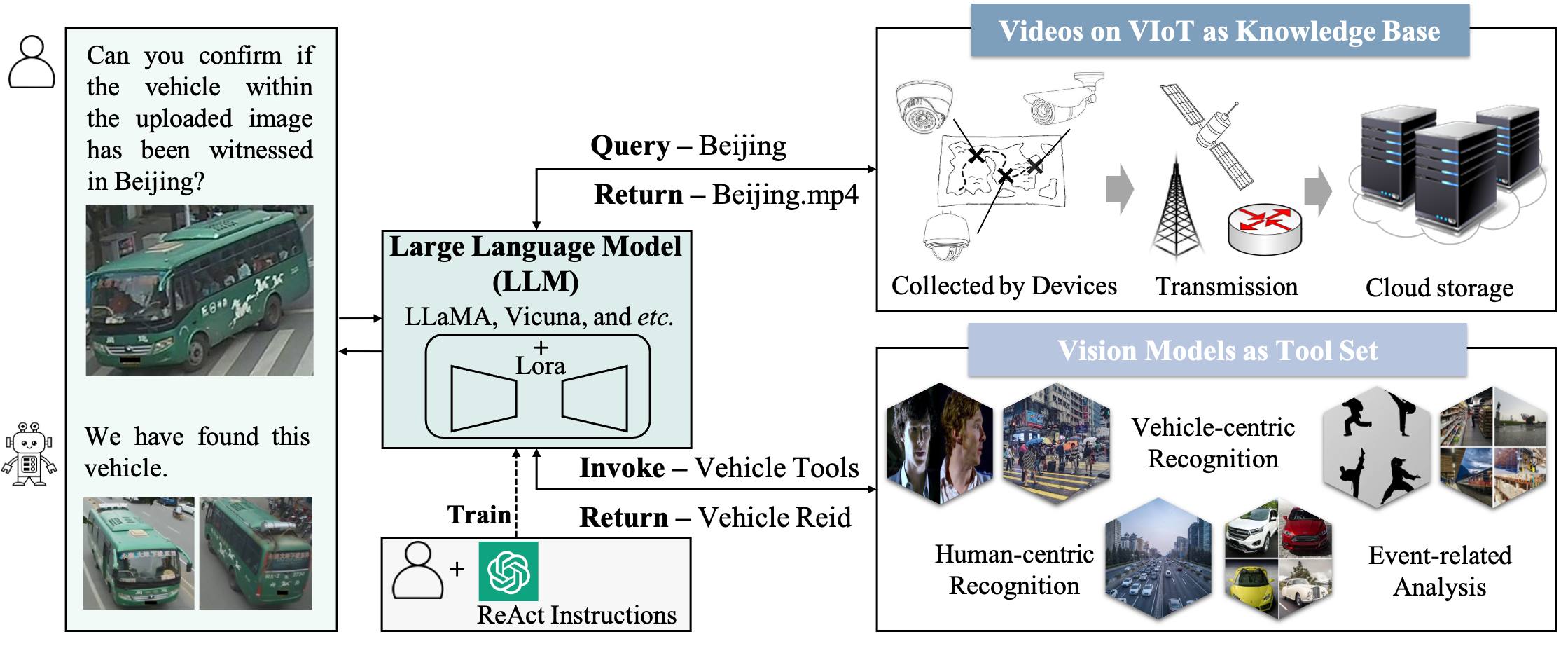}
    \caption{Illustration of VIoTGPT, which mainly consists of three fundamental modules, the videos that contain real-world observations, the human-centric algorithms acting as the tool set, and LLM as the intelligent agent.}
\label{fig:frame}
\end{figure*}

The ubiquitous visual sensors, contemporary communication technologies, and high-capacity networking have enabled the potential and widely-usage of Video Internet of Things (VIoT), \ie, internetworking of large-scale visual sensors, in collecting the unprecedented volume of video data and therefore offering a full ambient environment monitoring~\cite{ma2011internet,Mohan2017internet,chen2020internet,10.1145/3581783.3613910}. 

Leveraging the power of perceiving techniques driven by deep learning~\cite{k2012imagenet,simonyan2014very,he2016deep,hu2017squeeze,Dosovitskiy2021An}, the acquired video data is eagerly expected to help with various VIoT applications, such as large geographical area monitoring for smart transportation~\cite{liu_large-scale_2016,zantalis2019review,wei2019autoencoder} and public safety~\cite{liu2019exploring,zhang2022influence,liu2020enhancing}. 
As summarized by previous works~\cite{chen2020internet,zhang2020empowering,10.1145/3581783.3613910}, perceiving techniques are highly varied and wide in terms of analysis targets, understanding granularities, practical applications, from biometric recognition~\cite{Sun2014Deep,guo2016msceleb,Liu2017SphereFace,deng2019arcface,zhong2021sface}, human performance analysis~\cite{ji20123d,caba2015activitynet,zhang2019view,qi2020few,yun2024weakly}, to generic scene understanding~\cite{GrantF2017Crowd,qi2020stc,lv2024disentangled}, \etc. Therefore, how to schedule these domain-specific perceiving models and analyze the collected videos uniformly, efficiently, and especially intelligently is a main technique challenge. 

Although the general-purpose visual models~\cite{kirillov2023segany,zhang2023recognize} are compelling, they may not possess the domain-specific knowledge needed to replace certain perceiving models, especially the fine-grained ones like biometric recognition. Additionally, they are still too heavy to deal with a large volume of video data. 
Some recent works~\cite{schick2023toolformer,yao2023react,yang2023gpt4tools,wu2023visual,hao2023toolkengpt,kim2023guide,qin2023tool} discover the potential ability of the large language models (LLMs) to act as an intelligent agent to use tools, which motivates us to investigate the power of LLMs invoking and scheduling a variety of lightweight visual models to analyze the diverse surveillance videos. 

One possible approach is to directly guide powerful LLMs to use the tool by offering concise explanations and demonstrations through in-context prompts, such as VISPROG~\cite{gupta2023visual}, ViperGPT~\cite{suris2023vipergpt} and Visual ChatGPT~\cite{wu2023visual}. However, these works must rely heavily on strong LLMs like ChatGPT. Alternatively, another approach is to fine-tune LLMs to become proficient in particular tools, such as Toolformer~\cite{schick2023toolformer} and GPT4Tools~\cite{yang2023gpt4tools}, requiring deep acquaintance with the application domains and tool usages.

Compared with previous tools like calculators, search engines, and usual vision models, \etc, the visual algorithms for intelligent surveillance can be more ``unusual'' for LLMs to distinguish, plan, and execute. These algorithms fall under three primary categories: human-centric, vehicle-centric, and event-related. 
On one hand, LLMs necessitate distinguishing some fine-grained visual tools, \emph{e.g}., decide to invoke face recognition~\cite{deng2019arcface}, person re-identification~\cite{zheng2018pedestrian} or gait recognition~\cite{fan2023opengait} algorithm when they are asked to recognize the person in the video. 
On the other hand, it is required to invoke multiple interrelated algorithms successively and decide whether to execute the next visual algorithms, \emph{e.g}., whether to evaluate the impact of behavior after the action recognition~\cite{zheng2018pedestrian}. 

To address the above challenges, we propose a hierarchical taxonomy of VIoT tools and meticulously crafted the VIoT-Tool dataset involving 11 types of representative vision tools across three primary categories. To lead the LLM to decide actions not only based on the initial human queries but also considering the observation from the environment, we use ReAct instructions while not only using in-context learning or chain-of-thought prompting. We resort to instruction tuning to guide the LLM to learn a variety of instructions to accomplish complicated tasks with a given request. Specifically, we design ReAct instructions based on these representative tools by semi-automatic annotations, and then supervised finetune LLMs to learn the fine-grained difference of similar tools and multi-step reasoning for interrelated tools. 

The established framework, named VIoTGPT, is illustrated in Figure~\ref{fig:frame}. VIoTGPT consists of three fundamental components, \ie, LLM as the agent, videos as the knowledge base, and vision models as the tool set. The three fundamental components fulfill their duties and work seamlessly to deliver practical and effective results. 
To provide both quantitative and qualitative analysis, we establish the corresponding dataset VIoT-Tool based on diverse data including publicly available, web-collected, or self-made surveillance videos. Our major contributions can be summarized as follows:
\begin{itemize}
	\item{We propose VIoTGPT, the framework that applies the customized LLM as the intelligent agent, to interact with videos collected on VIoT, invoke visual models according to queries, and reply to human users. As we know, VIoTGPT is the first intelligent agent system to invoke visual models and analyze video data for intelligent video surveillance.}
	\item{To enable VIoTGPT, we propose a hierarchical taxonomy of VIoT, develop the training dataset involving 11 types of representative vision tools across three primary categories, and build the corresponding benchmarks to evaluate the performance of the intelligent agents. The dataset, named VIoT-Tool, will be publicly available to promote further research.}
    \item{Demonstrated by experimental results on VIoT-Tool, with instruction tuning, VIoTGPT can schedule the domain-specific perceiving models and analyze the collected videos intelligently to accomplish complicated tasks, especially can learn the fine-grained and interrelated tools scheduling ability.}
\end{itemize}

\section{Related Work}
\label{sec:formatting}
\subsection{Foundation Model.} 
The artificial intelligence landscape has been dominated by task-specific deep models~\cite{Liu2017SphereFace,caba2015activitynet,GrantF2017Crowd}, while a new wave of foundation models aims to gain general-purpose vision representations~\cite{chen2021exploring,kirillov2023segany,zhang2023recognize}, multi-modal representations~\cite{beit3,girdhar2023imagebind,li_blip-2_2023}, and even general-purpose generative models~\cite{videoworldsimulators2024}. 

The Segment Anything Model (SAM)~\cite{kirillov2023segany}, a pixel-level pre-training foundation model, has received widespread attention for its impressive ability in image segmentation. ImageBind~\cite{girdhar2023imagebind} learns joint representation across six different modalities including image, text, audio, depth, thermal, and IMU data, and enables cross-modal retrieval, cross-modal generation, \etc. The amazing Sora model~\cite{videoworldsimulators2024}, a text-conditional diffusion model based on transformer architecture, can generate high-fidelity video up to a minute long. It is trained on videos and images of varying durations, resolutions, and aspect ratios. 

Nevertheless, current vision-centered foundation models still struggle to generalize across different vision tasks or different data domains~\cite{tang2023can,ji2023segment,zhou2023can}. The difficulties in multiple-granularity understanding, temporal analysis, and comprehensive data domains and modalities together impede the development of a unified model for all the tasks~\cite{li2023multimodal}. We are eager to see the development of future unified models to replace the task-specific vision models of VIoT, then there is no need to invoke many fragmentary models. However, until then, we have to resort to other technical approaches for intelligence scheduling. We propose to construct a hierarchical taxonomy of the task-specific vision models and use the vision models as tools to collaboratively serve VIoT applications.

\subsection{LLMs and Tool learning.} 
Recently, large language models (LLMs) have received widespread attention because of their impressive performance on complex natural language understanding tasks~\cite{zhao2023survey}. LLMs are Transformer language models with billions of parameters trained on massive amounts of text data, which lead to some particularly interesting emergent behaviors including in-context learning~\cite{brown2020language}, instruction following~\cite{wei2022finetuned}, and step-by-step reasoning~\cite{wei2022chain}. Notable LLMs contains OpenAI's GPT-series~\cite{brown2020language} used in ChatGPT, PaLM of Google~\cite{chowdhery2022palm}, LLaMa of Meta~\cite{touvron2023llama,touvron2023llama2}, Vicuna~\cite{vicuna2023}, ChatGLM~\cite{du2022glm}, \etc. 

Despite the deficiencies such as limited corpora knowledge and unsatisfied numerical computation ability~\cite{zhao2023survey}, recent research has unveiled the potential of LLMs in mastering tools~\cite{qin_tool_2023,schick2023toolformer,shen2023hugginggpt,gupta2023visual,suris2023vipergpt,yang2023mm,wu2023visual,yang2023gpt4tools,gao2024clova}, enabling them to acquire domain-specific expertise and external knowledge. 
Without the need for explicit training, it is possible to accomplish tasks solely relying on the in-context learning ability of LLMs. VISPROG~\cite{gupta2023visual} and ViperGPT~\cite{suris2023vipergpt} have demonstrated that, with a small number of \emph{in-context examples instructions}, a powerful LLM GPT-3 can generate Python-like modular programs and execute them to invoke vision models and other functions for compositional visual tasks. 
ReAct~\cite{yao2023react} improved prompt engineering by introducing collaborative reasoning and execution, incorporating additional information into inference, and facilitating interaction between LLMs and the external environment. 
Based on ReAct and the powerful ChatGPT, MM-REACT~\cite{yang2023mm} and Visual ChatGPT~\cite{wu2023visual} can integrate the system with vision models using \emph{zero-shot prompting}, to accomplish visual understanding and generation tasks by invoking vision models and receiving the feedback iteratively until reaches the ending condition. CLOVA~\cite{gao2024clova} incorporates both correct and incorrect examples for prompts to generate better plans and programs, and design the reflection and learning scheme to update tools. 
Without the strong LLMs like GPT-3 and ChatGPT, some recent works~\cite{schick2023toolformer,yang2023gpt4tools,qin_tool_2023} explore the tool learning based on LLMs with about tens of billions of parameters(\eg GPT-J~\cite{gpt-j}, LLaMA 7B~\cite{touvron2023llama} and Vicuna 13B~\cite{vicuna2023}), with self-instruction tuning~\cite{wang2022self}, where instructions generated based on ChatGPT are used to fine-tune LLMs with a limited set of tools and have demonstrated promising results. 

Considering the limited computing resources, we also investigate how to leverage LLMs with tens of billions parameters effectively. While different with~\cite{schick2023toolformer,yang2023gpt4tools}, we mainly focus on learning to schedule vision tools and query knowledge videos towards intelligent Video Internet of Things (VIoT), and use instruction tuning of semi-automatic annotated ReAct instructions to keep the diversity and correctness of instructions. 

\begin{figure*}[htbp]
	\center
        \includegraphics[width=1\linewidth]{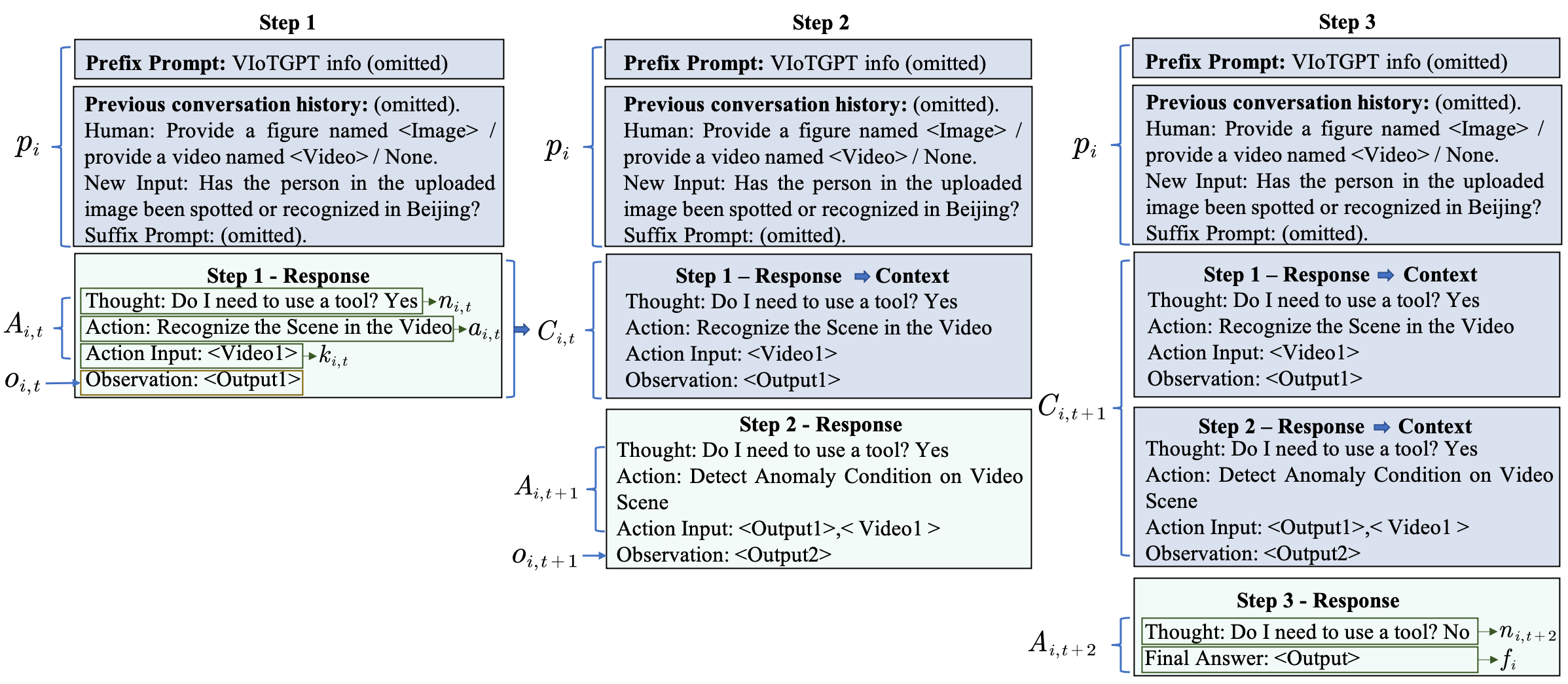}
	\caption{Pre-defined instructions and the response $A_{i,t}$ of the LLM $\theta$ at each step. In the intermediate steps, the LLM decides to use the tools (``Thought''), selected tool names (``Action''), and the input of tools  (``Action Input''). At the end of each step, the context will be updated by combing previous actions and observations to conversation history $C_{i,t} = (C_{i,t-1}, A_{i,t}, o_{i,t})$. At the final step, the LLM decides not to use tools and returns the final feedback $fi$.}
	\label{fig:template}
\end{figure*}

\section{VIoTGPT}
\subsection{Overview}
VIoTGPT consists of three components that collaborate to ensure practicality, as shown in Figure~\ref{fig:frame}.

\textbf{Video Data as Knowledge Base.} In practical VIoT applications, videos at various locations are collected by a variety of smart devices in real time. The massive scale videos contain real-world observations and can serve as the information-rich knowledge base denoted as $\mathcal{K} =\left\{ K_1,K_2,\cdots ,K_n \right\}$, which provides a solid foundation for VIoTGPT. For convenience, we use a knowledge base of videos with city-level locations, which can also be extended to more various kinds of tags, such as accurate spatio-temporal information, environmental information, \etc. 

\textbf{Perceiving Models as Tool Set.} We propose a hierarchical taxonomy of VIoT tools. The perceiving algorithms for VIoT can be concluded as mainly three primary categories: human-centric, vehicle-centric, and event-related algorithms. Each primary category can conclude a variety of secondary categories. Without loss of generality, we define the secondary categories by 11 representative vision algorithms including face recognition, person re-identification, gait recognition, vehicle re-identification, license plate recognition, human pose estimation, human action recognition, crowd counting, scene recognition, fire and smoke detection, and violence detection, as the toolset denoted as $\mathcal{T} =\left\{T_1,T_2,\cdots,T_m \right\}$, to demonstrate the feasibility and functionality of VIoTGPT. The agent can invoke one or several tools intelligently to accomplish the tasks, which will be detailed further in Section~\ref{sec:exp_setting}. 

\textbf{LLM as Agent.} The large language model (LLM) with model parameter $\theta$ is adopted as the intelligent agent to interact with humans, by firstly summarizing the input instructions and visual information to the pre-defined template, planning, observing, and reasoning with the assistance of the knowledge base and the tool set, and finally providing users with integrated processing information in its reply. Despite the powerful reasoning ability of LLMs such as LLaMa~\cite{touvron2023llama,touvron2023llama2} and GPT4Tools~\cite{yang2023gpt4tools}, it is still difficult for them to be directly used by prompting in VIoTGPT. To help LLMs become more proficient in querying knowledge and invoking tools, we finetune them with the meticulously crafted VIoT-Tool dataset.

\subsection{Instructions and Training}
\textbf{ReAct Instruction.} Considering the intelligent agent should interact with humans, tools, video data, and the environment, we introduce ReAct instruction for VIoTGPT to determine the action in each step not only based on the context but also considering the observation from outputs of previously invoked tools. 

Specifically, given the human query $q_i$ with potential visual information $v_i$, the LLM will summarize and format them and the overall framework information ($\mathcal{T}$ and $\mathcal{K}$) into a new prompt $p_i=\left( \mathcal{T},\mathcal{K},q_i,v_i \right)$ with the pre-defined template following Langchain~\cite{yao2023react,chase2022langchain}, as shown in Figure~\ref{fig:template}. 
With the input prompt $p_i$, at each step, the LLM will determine the action $A_{i,t}$, record observations of tools $o_{i,t}$, memorize the context $C_{i,t}=\left( A_{i,1},o_{i,1},\cdots, A_{i,t},o_{i,t} \right) $ (including the history of actions and observations), and reason iteratively until it achieves the final answer $f_{i}$. 

The action of the LLM ($\theta$) at step $t$ is represented by
\begin{equation}
    A_{i,t}=\begin{cases}
	\left( n_{i,t},a_{i,t},k_{i,t} \right) ,&		t\in \left\{ 0,1,\cdots ,T-1 \right\}\\
	\left( n_{i,t},f_i \right) ,&		t=T\\
\end{cases}
\end{equation}and determined by 
\begin{equation}
   P_{\theta}\left( A_{i,t} \right) =P_{\theta}\left( A_{i,t}|p_i,C_{i,t-1}\right), 
\end{equation}where $n_{i,t}$ denotes the \emph{decision} of whether to use tools $\mathcal{T}$. Ideally, in the intermediate steps, $n_{i,t}$ represents that the LLM determines to use tools (see $n_{i,t}$ in Figure~\ref{fig:template}), $a_{i,t}$ represents the selected \emph{tool names}, and $k_{i,t}$ represents the \emph{input} information (usually queried from $\mathcal{K}$) of the selected tool $a_{i,t}$. At the end of each step, the context will be updated $C_{i,t} = (C_{i,t-1}, A_{i,t}, o_{i,t})$ to record the history of actions and observations. At the final step $t=T$, $n_{i,t}$ represents not to use tools, the LLM will return the final feedback $fi$. 

To help the LLM $\theta$ perform in this way, we collect the ReAct instruction data $A_{i,t}$ and fine-tune the model. The ReAct instruction data is generated by ChatGPT and human annotations to keep the diversity of instructions and the correctness of tools and knowledge usages.

\begin{table}[]
\centering
\resizebox{\columnwidth}{!}{%
\begin{tabular}{@{}lll@{}}
\toprule[1.5pt]
Primary  Categories                            & Secondary  Categories     & Human Input       \\ \midrule
\multirow{4}{*}{Human-centric Recognition}     & Face Recognition          & Image,   Question \\ \cmidrule(l){2-3} 
 & Person Re-identification  & Image,   Question         \\ \cmidrule(l){2-3} 
 & Gait Recognition          & Video,   Question         \\ \cmidrule(l){2-3} 
 & Crowd Counting            & Question                  \\ \midrule
\multirow{2}{*}{Vehicle-centric   Recognition} & Vehicle Re-identification & Image,   Question \\ \cmidrule(l){2-3} 
 & License Plate Recognition & Question                  \\ \midrule
\multirow{5}{*}{Event-related Analysis}        & Fire Smoke Detection      & Question          \\ \cmidrule(l){2-3} 
 & Scene Recognition         & \multirow{2}{*}{Question} \\
 & Anomaly Recognition       &                           \\ \cmidrule(l){2-3} 
 & Pose Estimation           & \multirow{2}{*}{Question} \\
 & Action Recognition        &                           \\ \bottomrule[1.5pt]
\end{tabular}%
}
\caption{Summarization of VIoT-Tool.}
\label{fig:tools}
\end{table}

\begin{table*}[htbp]
\centering
\resizebox{\textwidth}{!}{%
\begin{tabular}{@{}lccccccccccccc@{}}
\toprule[1.5pt]
\multirow{2}{*}{Models} &
  \multirow{2}{*}{Promt} &
  \multicolumn{4}{c}{Single Tool Responses} &
  \multicolumn{4}{c}{Interrelated Tools Responses} &
  \multicolumn{4}{c}{All Responses} \\ \cmidrule(l){3-14} 
& & Decis & Tool  & Input & Whole & Decis & Tool  & Input & Whole & Decis & Tool  & Input & Whole \\ \midrule
Llama   & Zero-shot                    
& 0.00  & 0.00  & 0.00  & 0.00  
& 0.00  & 0.00  & 0.00  & 0.00  
& 0.00  & 0.00  & 0.00  & 0.00  \\

Llama   & Few-shot                    
& 0.00  & 0.00  & 0.00  & 0.00  
& 0.00  & 0.00  & 0.00  & 0.00  
& 0.00  & 0.00  & 0.00  & 0.00  \\

\rowcolor{gray!20} VIoTGPT (Llama) &
  Zero-shot &
  \textbf{89.44} &
  \textbf{41.85} &
  \textbf{75.53} &
  \textbf{41.11} &
  \textbf{84.79} &
  \textbf{69.75} &
  \textbf{69.13} &
  \textbf{67.25} &
  \textbf{87.68} &
  \textbf{48.35} &
  \textbf{74.04} &
  \textbf{47.20} \\ \midrule
Vicuna  & Zero-shot  
& 86.39 & 60.64 & 0.00 & 0.00  
& 47.38 & 8.00  & 0.00  & 0.00  
& 71.64 & 48.37 & 0.00  & 0.00  \\

Vicuna  & Few-shot  
& 28.88 & 26.98 & 29.03 & 22.34 
& 1.00 & 0.00 & 0.00 & 0.00 
& 22.33 & 20.69 & 22.26  & 17.13  \\

\rowcolor{gray!20} VIoTGPT (Vicuna) &
  Zero-shot &
  \textbf{99.01} &
  \textbf{63.83} &
  \textbf{76.06} &
  \textbf{35.79} &
  \textbf{86.11} &
  \textbf{74.38} &
  \textbf{69.75} &
  \textbf{52.00} &
  \textbf{94.13} &
  \textbf{66.29} &
  \textbf{74.59} &
  \textbf{39.59} \\ \midrule
ChatGPT & Zero-shot  
& 99.92 & 96.80 & 96.73 & 93.92 
& 74.16 & 19.75 & 15.50 & 0.00  
& 89.53 & 78.85 & 77.80 & 72.03 \\
ChatGPT & \multicolumn{1}{l}{Few-shot} 
& 99.85 & 99.85 & 99.85 & 99.85 
& 98.57 & 95.38 & 61.63 & 28.25 
& 99.24 & 98.81 & 90.94 & 83.16 \\ \bottomrule[1.5pt]
\end{tabular}%
}
\caption{Quantitative results on VIoT-Tool. ``Decis'' represents the accuracy of decisions of whether to use tools $Acc_{n_{i,t}}$. ``Tool'' represents the accuracy of chosen tool names $Acc_{a_{i,t}}$. ``Input'' represents the accuracy of input information of tools $Acc_{k_{i,t}}$. ``Whole'' represents the accuracy of the whole response $Acc_{A_{i,t}}$. ``IT'' represents instruction tuning. }
\label{table:main}
\end{table*}

\textbf{Supervised Fine-tuning.}
The learning process of the LLM $\theta$ can be formulated as:  
\begin{equation}
L_{sft}\left( \theta \right) =-\sum_i{\sum_t{ \log P_{\theta}\left( A_{i,t}|p_i,C_{i,t-1} \right).\\
\\
}}
\end{equation}With the supervised fine-tuning with instructions, LLM can invoke tools and query requisite knowledge of VIoT to perform targeted instructions, particularly in fine-grained tool usage and multi-step reasoning for interrelated tools.

\subsection{Tools, Training Dataset, and Benchmarks}
VIoT-Tool is established based on 11 represented vision tools, as shown in Table~\ref{fig:tools}
Considering privacy and copyright, images and videos used in this paper are publicly available~\cite{zheng_scalable_2015}, web-collected~\cite{Nagrani17b}, or self-made surveillance videos. 
For convenience, all the videos used in the knowledge base are represented as 125 city-level names. 


\begin{figure*}[htbp]
	\center
        \includegraphics[width=0.98\linewidth]{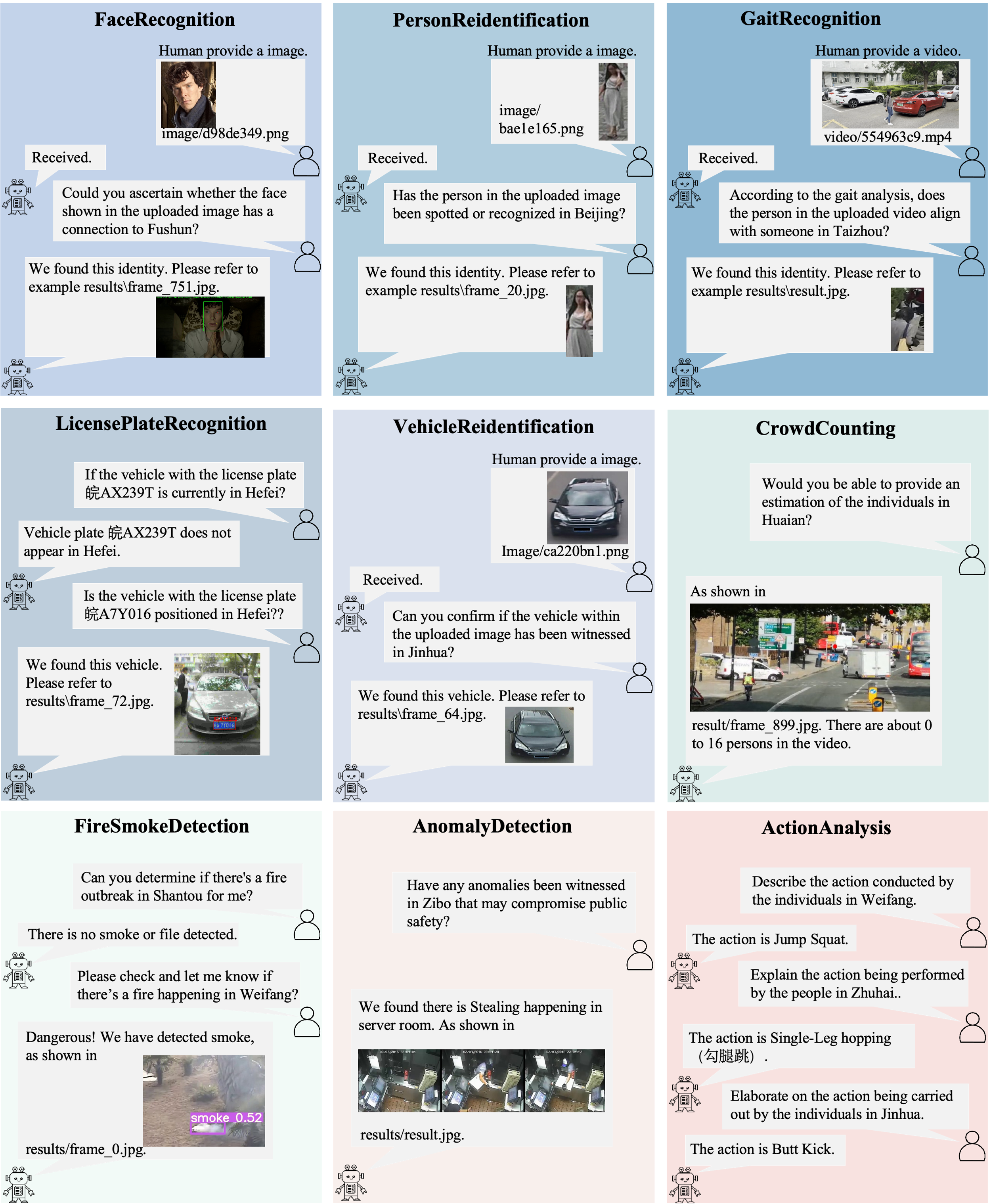}

	\caption{Illustration of VIoTGPT's capabilities and applications. ``FaceRecognition'', ``PersonReidentification'', ``GaitRecognition'', ``LicensePlateRecognition'', ``VehicleReidentification'', ``CrowdCounting'' and ``FireSmokeDetection'' represent responses with single tool. While ``AnomalyDetection'' and ``ActionAnalysis'' are two event-related pipelines that require scheduling interrelated tools. }
	\label{fig:result}
\end{figure*}

\subsubsection{Dataset Summarization} 
\
\newline
\indent\textbf{Statistics.} Through meticulous annotation and optimization, we collect the training dataset with training instructions (2.79 billion tokens) constructed by 200K pairs ($p_i$ and $A_{i,t}$) related to the 11 tools across three categories and the corresponding testing datasets with 1,841 pairs. To avoid data bias, the tool data distribution of the training and the test set are almost consistent, which is detailed in the Appendix. 

\textbf{Generalization and Robustness.} To evaluate the generalization of LLMs on the knowledge videos, the video data in the testing benchmarks is different from that in the training dataset. To evaluate the semantic robustness, the instructions in the testing datasets are quite different from those in the training dataset. 

\textbf{Fine-grained and Interrelated Tools.} In the hierarchical taxonomy of VIoT tools, tools of human-centric recognition or vehicle-centric recognition usually share similar objectives and usage, therefore leading to fine-grained tool differentiation. Event-related analysis typically involves multiple interrelated tools, making them useful as evaluation metrics for this purpose. 

\section{Experiments}
\label{sec:exp_setting}
\subsection{Experimental Setting}

\subsubsection{Models Details and Baselines.}
With limited computing resources, we mainly investigate how to leverage LLMs with tens of billions of parameters effectively. We use Llama 7B~\cite{touvron2023llama} and Vicuna 7B~\cite{vicuna2023} as base models for the following fine-tuning. Correspondingly, Llama 7B and Vicuna 7B without fine-tuning are used as baselines, which rely on the in-context ability with the same prompt. ChatGPT (gpt-3.5-turbo) is also used to compare our models and demonstrate an optimal performance on VIoT-Tool. 

\subsubsection{Training Details.} To enable training, a parameter-efficient tuning method, \emph{i.e.}, Low-Rank Adaptation (LoRA)~\cite{hu2022lora}, is used. Specifically, we attach the LoRA modules to the query and key of self-attention layers, with the rank parameter 8, the scaling alpha parameter 16, and the dropout rate 0.05, following the settings of FastChat~\cite{zheng2023judging}. The maximum length of new tokens is 2,048. We finetune LLMs using an effective batch size of 256 and a learning rate of 5e-5 for 6 epochs with the AdamW optimizer. In the training process, the instruction datasets are randomly divided into training and evaluating sets in a 49:1 proportion. All the experiments are conducted on NVIDIA RTX 4090 GPUs.

\subsubsection{Evaluation Metrics.}
Four main evaluation metrics are used to evaluate the performance, including accuracy of decisions, accuracy of tool names, accuracy of tool inputs, and accuracy of whole response. Specifically, $n_{i,t}$ represents the decision of whether to use tools (``Thought: Do I need to use a tool? Yes/No''), and $Acc_{n_{i,t}}$ measures accuracy of $n_{i,t}$. Then the accuracy of decisions is defined as
\begin{equation}
Acc_{n_{i,t}}=\frac{1}{\sum_i{\sum_t{}}}
 \sum_i{\sum_t{\{
	n_{i,t}=\\
}}\dot{n}_{i,t}\}
,\end{equation}where $\dot{n}_{i,t}$ represents the gold label of the decision whether to use tools at step $t$. Note that, $a_{i,t}$ represents the chosen tool name, and $Acc_{a_{i,t}}$ measures the accuracy of all the chosen tools. Then, the accuracy of tool names is
\begin{equation}
Acc_{a_{i,t}}=\frac{1}{\sum_i{\sum_t{}}}
\sum_i{\sum_t{\{
	a_{i,t}=\\}}\dot{a}_{i,t}\}
,\end{equation}where $\dot{a}_{i,t}$ represents the gold label of chosen tool at step $t$.
In addition, $k_{i,t}$ represents the input of tool $a_{i,t}$ including the the queried knowledge information. $Acc_{k_{i,t}}$ measures the accuracy of the input information. Then the accuracy of tool inputs is 
\begin{equation}
Acc_{k_{i,t}}=\frac{1}{\sum_i{\sum_t{}}}
\sum_i{\sum_t{\{
	k_{i,t}=\\}}\dot{k}_{i,t}\}
,\end{equation}where $\dot{k}_{i,t}$ represents the gold label of the input at step $t$.
Finally, $A_{i,t}$ represents the whole response of LLM, and $Acc_{A_{i,t}}$ measures its accuracy. The accuracy of the whole response is
\begin{equation}
Acc_{A_{i,t}}=\frac{1}{\sum_i}
\sum_i{(\prod_t{\{ A_{i,t}=\dot{A}_{i,t} \}})}
,\end{equation}where $A_{i,t}$ represents the gold label of response at step $t$.

\begin{table}[htbp]
\centering
\resizebox{\columnwidth}{!}{%
\begin{tabular}{@{}lcccccccc@{}}
\toprule[1.5pt]
\multirow{2}{*}{Models} & \multicolumn{4}{c}{Validation Set}                                & \multicolumn{4}{c}{Test Set}                                      \\ \cmidrule(l){2-9} 
& Decis          & Tool           & Input          & Whole          & Decis          & Tool           & Input          & Whole          \\ \midrule
Llama                   
& 0.00           & 0.00           & 0.00           & 0.00           
& 0.00           & 0.00           & 0.00           & 0.00           \\

\rowcolor{gray!20}  VIoTGPT (Llama)              
& \textbf{84.05} & \textbf{61.96} & \textbf{81.60} & \textbf{60.13} 
& \textbf{89.44} & \textbf{41.85} & \textbf{75.53} & \textbf{41.11} \\ \midrule

Vicuna                  
& 81.86          & 61.24          & 0.00          & 0.00           
& 86.39          & 60.64          & 0.00          & 0.00           \\

\rowcolor{gray!20}  VIoTGPT (Vicuna) 
& \textbf{98.16} & \textbf{66.26} & \textbf{76.07} & \textbf{58.44} 
& \textbf{99.01} & \textbf{63.83} & \textbf{76.06} & \textbf{35.79} \\ \bottomrule[1.5pt]
\end{tabular}%
}
\caption{Comparison of the single tool performance on validation and test sets of VIoT-Tool respectively.}
\label{table:com1}
\end{table}

\begin{table}[h]
\centering
\resizebox{\columnwidth}{!}{%
\begin{tabular}{@{}lcccccccc@{}}
\toprule[1.5pt]
\multirow{2}{*}{Models} & \multicolumn{4}{c}{Validation Set}                                & \multicolumn{4}{c}{Test Set}                                      \\ \cmidrule(l){2-9} 
& Decis          & Tool           & Input          & Whole          & Decis         & Tool           & Input          & Whole          \\ \midrule
Llama                   
& 0.00           & 0.00           & 0.00           & 0.00           
& 0.00           & 0.00           & 0.00           & 0.00           \\
\rowcolor{gray!20}  VIoTGPT (Llama)              
& \textbf{88.72} & \textbf{76.33} & \textbf{76.67} & \textbf{71.33} 
& \textbf{84.79} & \textbf{69.75} & \textbf{69.13} & \textbf{67.25} \\ \midrule
Vicuna                  
& 45.01          & 10.25          & 0.00           & 0.00           
& 47.38          & 8.00           & 0.00           & 0.00           \\
\rowcolor{gray!20}  VIoTGPT (Vicuna)
& \textbf{90.08} & \textbf{77.67} & \textbf{80.33} & \textbf{66.00} & \textbf{86.11} & \textbf{74.38} & \textbf{69.75} & \textbf{52.00} \\ \bottomrule[1.5pt]
\end{tabular}%
}
\caption{Comparison of interrelated tools performance on validation and test sets of VIoT-Tool.}

\label{table:com2}
\end{table}

\begin{table}[h]
\centering
\resizebox{\columnwidth}{!}{%
\begin{tabular}{@{}lcccccccc@{}}
\toprule[1.5pt]
\multirow{2}{*}{Models} & \multicolumn{4}{c}{Validation Set}   & \multicolumn{4}{c}{Test Set} \\ \cmidrule(l){2-9} 
& Decis          & Tool           & Input          & Whole     & Decis          & Tool           & Input          & Whole     \\ \midrule
Llama                   
& 0.00           & 0.00           & 0.00           & 0.00           
& 0.00           & 0.00           & 0.00           & 0.00           \\
\rowcolor{gray!20}  VIoTGPT (Llama)              
& \textbf{85.82} & \textbf{65.31} & \textbf{80.45} & \textbf{62.74} 
& \textbf{87.68} & \textbf{48.35} & \textbf{74.04} & \textbf{47.20} \\ \midrule
Vicuna                  
& 67.93 & 49.35 & 0.00  & 0.00           
& 71.64 & 48.37 & 0.00  & 0.00 \\
\rowcolor{gray!20}  VIoTGPT (Vicuna)              
& \textbf{95.11} & \textbf{68.92} & \textbf{77.06} & \textbf{60.20} 
& \textbf{94.13} & \textbf{66.29} & \textbf{74.59} & \textbf{39.59} \\ \bottomrule[1.5pt]
\end{tabular}%
}
\caption{Comparison of tools performance on validation and test sets of VIoT-Tool.}
\label{table:com3}
\end{table}

\subsection{Experimental Results}
\subsubsection{Main results.} 
With the instruction tuning, LLMs have gained the ability to invoke tools and unseen videos given the unseen human queries. Table~\ref{table:main} and Figure~\ref{fig:result} report the main experimental results. The shown instructions and responses are in accordance with Table~\ref{fig:tools}. 

\subsubsection{Observations and Discussions} 
\
\newline
\indent\textbf{Baselines.} We found that Llama 7B without fine-tuning could not follow the format requirements with the given prompt, \eg ``Thought: Do I need to use a tool? Yes'', let alone use the names of the tools correctly. 
Vicuna 7B without fine-tuning performs better than Llama since it can make correct decisions using ``Thought: Do I need to use a tool? Yes'' and even choose appropriate tools, but still cannot correctly query videos in the datasets and failed to accomplish the whole process. In addition, the generated sentences of original Vicuna 7B always contain extra whitespaces and they will lead to adverse effects (that we have managed to overcome to calculate accuracies in Table~\ref{table:main}). 
We also apply the Vicuna 13B model trained on GPT4Tools~\cite{yang2023gpt4tools} as the agent for VIoT. However, it has been observed that this model yields no significant gains compared to the original Vicuna model on our benchmarks. We speculate this is due to the different trained tools and the video querying requirements of VIoT. 

\textbf{VIoTGPT.} VIoTGPT models (Llama or Vicuna with instruction tuning on VIoT dataset, denoted by ``Llama+IT'' and ``Vicuna+IT'') have achieved significant performance improvement on both single tool responses and interrelated tools responses compared with Llama 7B and Vicuna 7B without any fine-tuning, as shown in Table~\ref{table:main}. 
For ``Llama+IT'', we observed that the accuracy of whole response $Acc_{A_{i,t}}$ of single tool responses is mainly limited by the accuracy of tool names $a_{i,t}$, especially some fine-grained differences, \eg distinguishing license plate recognition and vehicle re-identification, distinguishing gait recognition and person re-identification. 
This limitation also hinders the performance of ``Vicuna+IT''. Note that, the fine-grained difference between tools is not that challenging when LLMs only need to deal with single-tool responses. However, it is difficult to learn the fine-grained difference along with multi-step reasoning for interrelated tools simultaneously.  
For ``Vicuna+IT'', another interesting observation is that, although it has invoked tools and querying video knowledge correctly, it failed to return the final feedback (just giving nothing back) to finish the whole process $A_{i,t}$. 

\textbf{ChatGPT.} Not surprisingly, with powerful reasoning ability and in-context prompts, ChatGPT showed strong performance on the individual tool responses in the VIoT-Tool dataset, but it still exhibited significant errors with the interrelated tools, such as invoking tools repeatedly, using the wrong tools and the wrong input, and failing to give final responses due to hallucination. To make full use of the ability of ChatGPT, in addition to the zero-shot prefix prompt that only describes the function of the toolset, the few-shot prompt setting is also used to learn responses for specific human queries to help ChatGPT know how to use the tool set and the knowledge videos. ChatGPT (few-shot) can achieve a nearly satisfactory performance on single-tool responses. However, challenges remain in interrelated tools (even worse than VIoTGPT). Despite this, the performance still demonstrated the reliability of the VIoT-Tool dataset and showed the excellent reasoning ability of ChatGPT, which is what we pursue in the closed system with small-size agents without external APIs. 

\textbf{Additional Experimental Results.} 
We present a comparison of the performance of the validation set and the test set. The experimental results of single tool responses, interrelated tools responses, and all responses are listed in Table~\ref{table:com1}, Table~\ref{table:com2}, and Table~\ref{table:com3}. We found that there is still scope for improvement on the validation set. The experimental evidence suggests that the implementation of more powerful LLMs is imperative to achieve better results. In addition, there exists a significant performance gap between the validation set and the test set. It is evident that the performance gap is primarily due to the semantic differences in human queries or the diversity of queried knowledge videos. By identifying and addressing these challenges in the future, VIoTGPT can achieve better performance. 

\section{Conclusion}
In this paper, we presented VIoTGPT, a framework that trains LLMs as intelligent agents to interact with humans, query knowledge videos, and activate vision models to complete complex tasks for intelligent VIoT. We carefully constructed VIoT-Tool dataset to support VIoTGPT and future research. Based on these, we observed promising results. We believe scheduling perceiving models and analyzing videos intelligently will lead VIoT into a bright and smart future. 

\appendix
\setcounter{table}{0}
\setcounter{figure}{0}

\section{Appendix - Details of Tools, Training Dataset, and Benchmarks of VIoT-Tool}

VIoT-Tool is established based on 11 represented vision tools, as listed in Table 1 of the main paper. Specifically, tools and the corresponding training datasets and benchmarks are introduced as follows respectively. 

\subsubsection{1. Human-centric Tools}
\
\newline
\indent\textbf{Face Recognition.}
Face recognition algorithms~\cite{wang2021deep} recognize the identity of the given image from the candidate image or video gallery based on facial features. We use the OpenCV lightweight pipeline including YuNet~\cite{wu2023yunet} for detection and SFace~\cite{zhong2021sface} for recognition. To build the dataset, we collect 101 celebrity images from the internet as queries and select 70 video clips from the Sherlock~\cite{Nagrani17b} as the gallery. 

\textbf{Person Re-identification.} 
Person re-identification involves recognizing individuals across different cameras or scenes. This is valuable for video surveillance, where identifying individuals as they move through a network of cameras can enhance security and public safety. We use Fast-ReID~\cite{he_fastreid_2020} as the pedestrian re-identification model and collect videos from the Market-1501~\cite{zheng_scalable_2015} as the testing benchmark. 

\textbf{Gait Recognition.}
Gait recognition algorithm identifies people by analyzing their walking posture and comparing it to candidate videos. For the vision model, we choose the model in All-in-One-Gait, which is a subproject of OpenGait~\cite{Fan_2023_CVPR} built mainly based on ByteTrack~\cite{Zhang2022ByteTrack} and PaddleSeg~\cite{liu2021paddleseg}. We select a subset of videos from the dataset CASIA-B~\cite{szhengICIP2011} and combine them with some self-recorded videos as the testing benchmark.

\textbf{Crowd Counting.}
Crowd counting is to count people from candidate images or video frames of dense crowds. For the choice of model, the Point to Point Network (P2PNet) based on a purely point-based framework~\cite{song2021rethinking} is adopted to locate and count people in a crowd. We select video frames from the VSCrowd dataset~\cite{li2022video} and the Fudan-ShanghaiTech (FDST) dataset~\cite{fang2019locality} and synthesize them as the testing benchmark.

\subsubsection{2. Vehicle-centric Tools} 
\
\newline
\indent\textbf{Vehicle Re-identification.} Similar to pedestrian re-identification, vehicle re-identification~\cite{liu2016deep,liu2017provid,wang2020global} is used to recognize and match vehicles among cameras.  
This technology is widely employed in traffic management, surveillance, and security systems to monitor vehicles as they traverse various points within a network of cameras. 
We introduce a vehicle re-identification model and select a subset of the VeRi dataset~\cite{liu_large-scale_2016} as our testing benchmark.

\textbf{License Plate Recognition.}
License plate recognition involves the detection and identification of license plates on vehicles. We introduce a license plate detection model, which detects license plates within images and identifies the corresponding vehicle in the videos. We select video frames from the Chinese city parking dataset (CCPD)~\cite{ferrari_towards_2018} and synthesize them as the testing benchmark.

\subsubsection{3. Event-related Tools} 
\
\newline
\indent\textbf{Fire Smoke Detection.} 
Fire smoke detection is an algorithm that identifies and detects hazardous fire and smoke in videos to promptly eliminate fire hazards. In the paper, we use the lightweight Fire and Smoke Detection~\cite{geng2022fire} model based on Yolov4~\cite{bochkovskiy2020yolov4}. 70 in-the-wild videos have been collected from the internet to be used in the testing benchmark. 

\textbf{Pose Estimation.}
Pose Estimation detects the position and orientation of a person by predicting the location of specific keypoints like hands, head, elbows, \etc. In the paper, we build this tool based on MMPose~\cite{mmpose2020} and use self-made videos as the test benchmarks. 

\textbf{Action Recognition.} 
The fundamental goal of human action recognition is to analyze a video to identify the actions taking place in the video. In this model, we use the skeleton-based action recognition model following MMAction2~\cite{2020mmaction2} and use self-made videos as the test benchmarks.

\textbf{Scene Recognition.}
Scene Recognition identifies the location or environment of a video based on the background information. The paper applies the pre-trained model based on Places365~\cite{zhou2017places}. We annotate a subset of videos from the dataset UCF-Crime~\cite{sultani2018real} as the testing benchmark. 

\textbf{Violence Detection.}
Video violence detection extracts frame features and uses them to identify violent behavior. We use the I3D model~\cite{carreira2017quo} pre-trained on Kinetics-400 for feature extraction and employ a local-global context aggregation~\cite{pu2023learning} to recognize violent behaviors using the I3D features. We selected a subset of data from the UCF-Crime dataset~\cite{sultani2018real} to build the testing benchmark.

\section{Appendix - Details of VIoT-Tool}
This section concludes some details that are not fully explained in the paper, including details of the training
dataset, statistics of the VIoT-Tool dataset, details of the data annotation process, and details of the instruction templates of VIoTGPT.

\subsection{Statistics of VIoT-Tool}
\begin{figure}[htbp]
\renewcommand{\thefigure}{S\arabic{figure}}
	\center
    \includegraphics[width=1\linewidth]{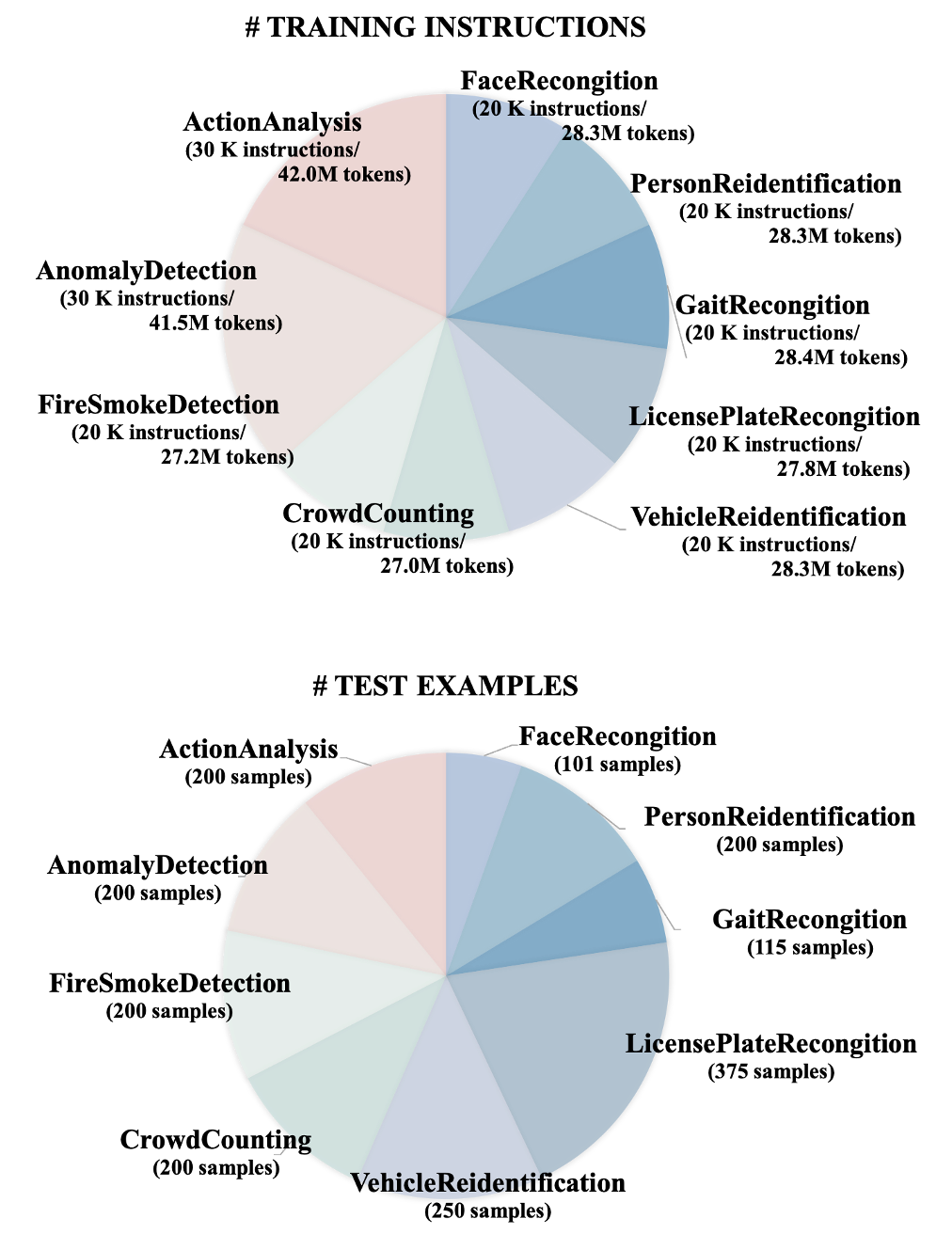}
	\caption{Statistics of training and testing dataset of VIoT-Tool. A total of 200K training instructions (2.79 billion tokens) and 1,841 test samples are concluded in the VIoT-Tool dataset, related to 11 tools across three categories.}
	\label{fig:dataset}
\end{figure}

There are 200K training instruction pairs related to the aforementioned 11 tools across three categories and the corresponding testing datasets with 1,841 samples in the VIoT-Tool dataset. The tool data distribution of the training set and the test set are almost consistent to avoid data bias, as shown in Figure~\ref{fig:dataset}. 

\subsection{Details of Data Annotation}
\begin{figure}[htbp]
\renewcommand{\thefigure}{S\arabic{figure}}
	\center
	\includegraphics[width=1\linewidth]{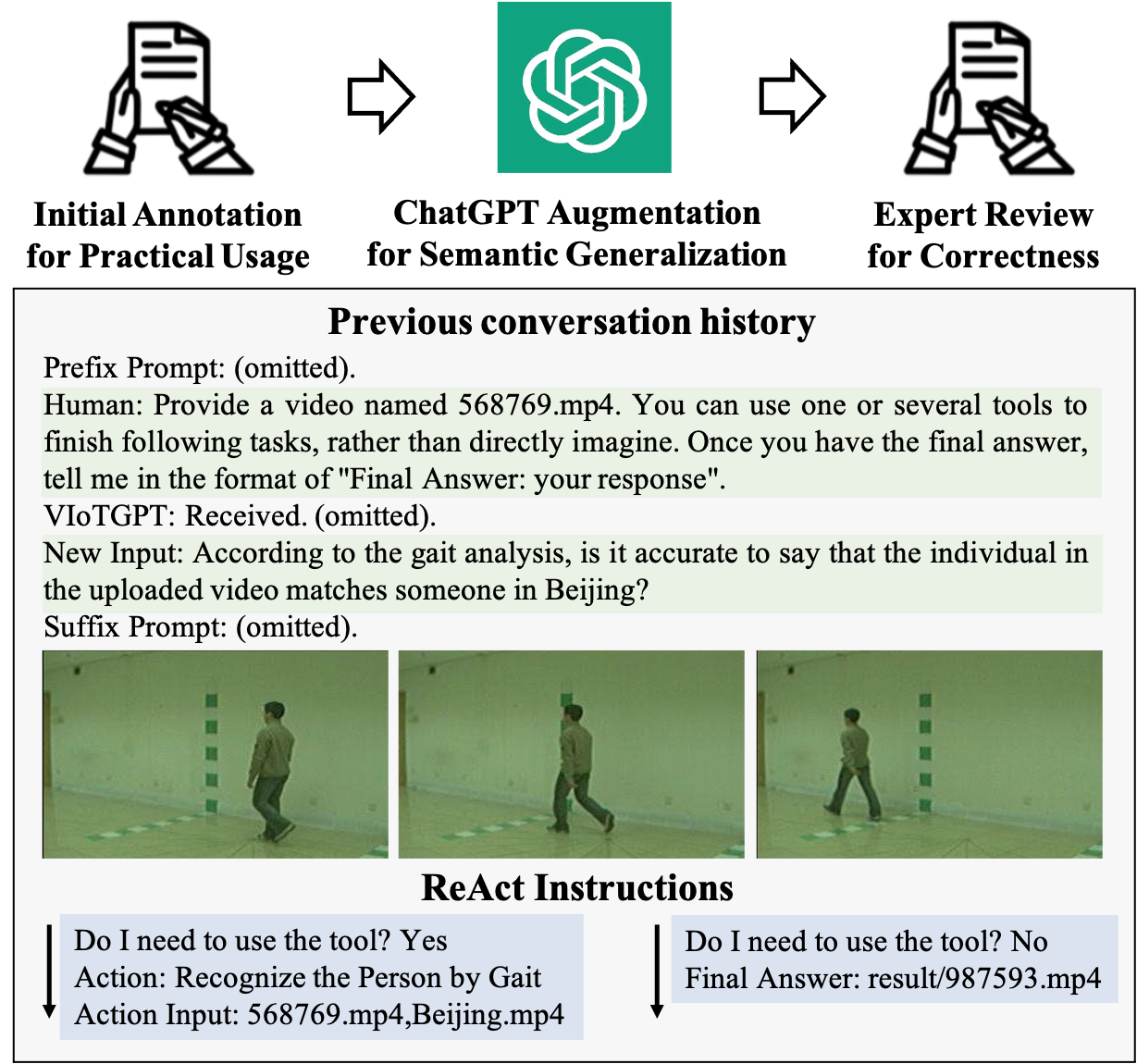}
	\caption{Illustration of semi-annotated instructions, with initial annotations for practicability, ChatGPT for semantic generalization, and expert review for correctness. }
	\label{fig:annotation}
\end{figure}

The instructions and golden labels are collected semi-automatically, with initial annotations for practicability, ChatGPT for semantic generalization, and expert review for correctness. Specifically, in the annotation process, the annotator can only know all information of the toolkit and will only use one specific location (\eg the city-level name like Beijing) to generate 
a few examples of human input $p_i$ (highlighted in green color in Figure~\ref{fig:annotation}) and the response $A_{i,t}$ (highlighted in blue color in Figure~\ref{fig:annotation}). ChatGPT can further augment the question of $p_i$ (``New Input'' highlighted in green in Figure~\ref{fig:annotation}) with different semantic variety and varied locations with the same $A_{i,t}$ to keep the data diversity. Finally, the expert will randomly review the $p_i$ and $A_{i,t}$ to keep the correctness. To evaluate the semantic robustness, questions of instructions $p_i$ in the testing datasets are ensured to be significantly different from those in the training dataset.

\section{Appendix - Details of Instruction Templates}
\begin{figure}[htbp]
\renewcommand{\thefigure}{S\arabic{figure}}
	\center
	\includegraphics[width=1\linewidth]{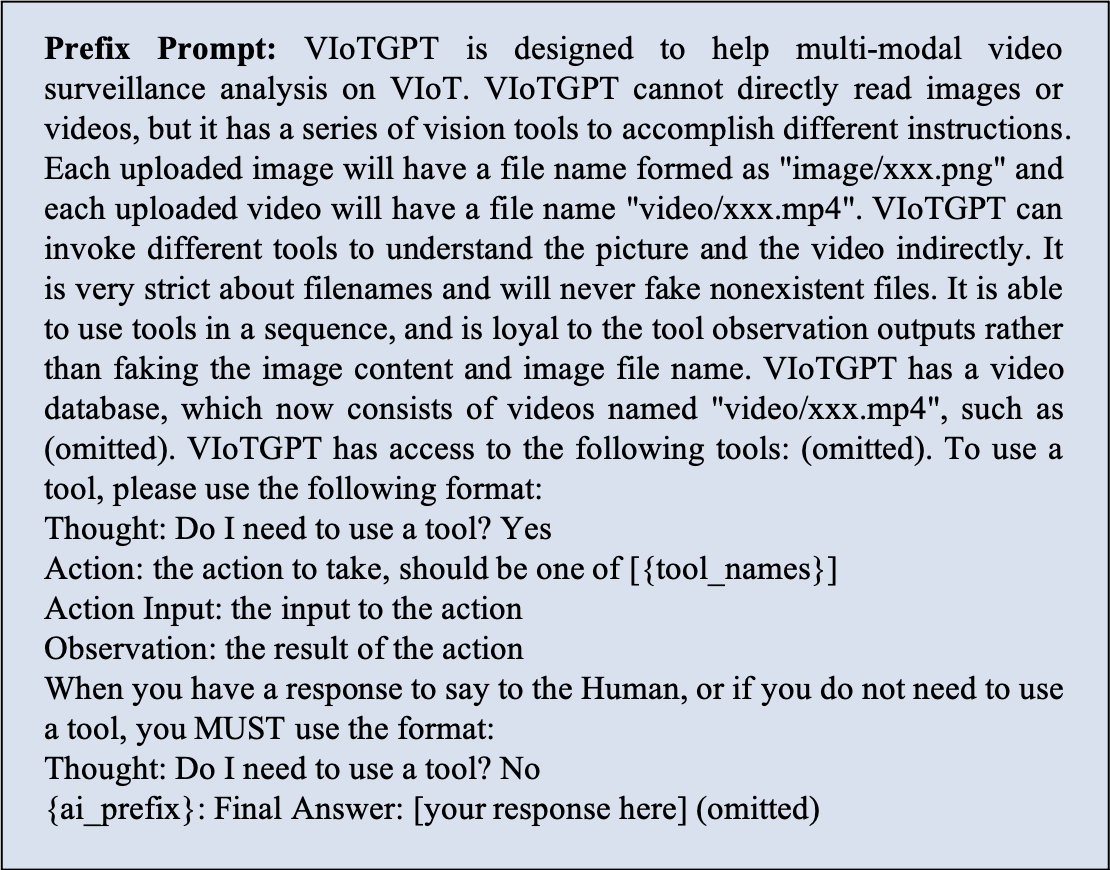}
	\caption{Prefix Prompt of instruction templates.}
	\label{fig:prefix}
\end{figure}

\begin{figure}[htbp]
\renewcommand{\thefigure}{S\arabic{figure}}
	\center
	\includegraphics[width=1\linewidth]{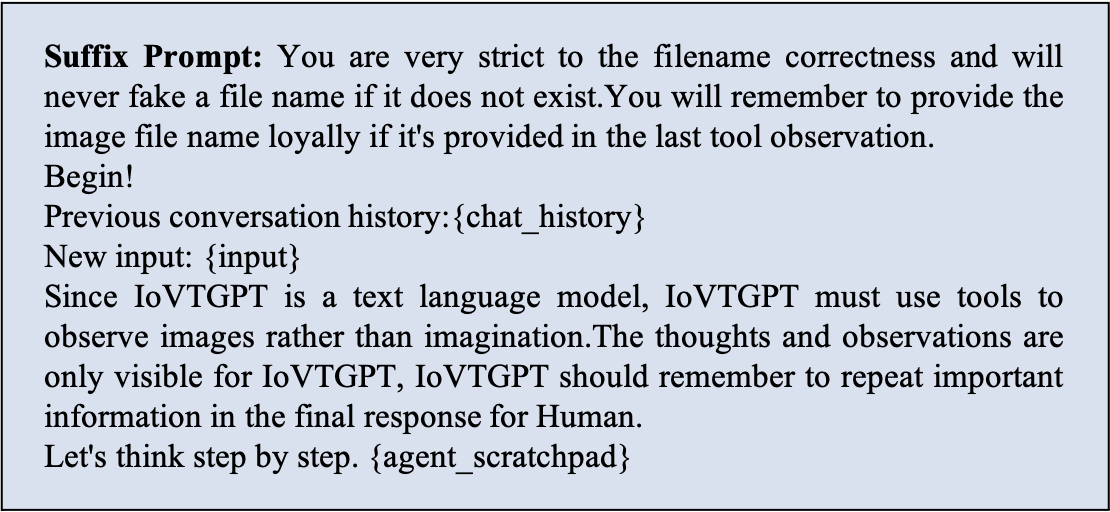}
	\caption{Suffix Prompt of instruction templates.}
	\label{fig:suffix}
\end{figure}

In the prompt engineering of VIoTGPT, we use ReAct~\cite{yao2023react} to introduce collaborative reasoning and execution, incorporate additional tool observations into inference, and facilitate interaction between LLMs and external tools. Specifically, the prefix prompt and suffix prompt are shown in Figure~\ref{fig:prefix} and Figure~\ref{fig:suffix}, respectively. The prefix prompt introduces the purpose, components, and usage of VIoTGPT. The prompt suffix incorporates previous conversation history and the new input (they are separated out from the suffix for easier understanding), and future incorporated context (``agent\_scratchpad''). 

\section{Appendix - Details of Experiments}
VIoTGPT models have achieved significant performance improvement on both single tool responses and interrelated tools responses, as listed in Table 2 of the main paper. 
For ``Llama+IT'', we observed that the accuracy of whole response $Acc_{A_{i,t}}$ of single tool responses is mainly limited by the accuracy of tool names $a_{i,t}$, especially some fine-grained differences, \eg distinguishing license plate recognition and vehicle re-identification, distinguishing gait recognition and person re-identification. 
The limitation also hinders the performance of ``Vicuna+IT''. Note that, the fine-grained difference between tools is not that challenging when LLMs only need to deal with single tool responses, as the comparison in Figure~\ref{fig:result2}. However, it is challenging to simultaneously learn the fine-grained difference along with multi-step reasoning for interrelated tools.  

\bibliography{aaai25}
\end{document}